\documentclass[]{stem}
\usepackage{microtype}
\usepackage{amsfonts}
\usepackage{graphicx}
\usepackage{booktabs}
\usepackage{wrapfig}
\usepackage{amsmath}
\usepackage{amsthm}
\usepackage{subcaption}
\usepackage{float}

\usepackage{algpseudocode}
\usepackage[linesnumbered,lined,boxed,commentsnumbered,ruled,longend]{algorithm2e}
\usepackage{mathtools}
\usepackage{arydshln}
\usepackage{seqsplit}
\usepackage{threeparttable}
\usepackage{tablefootnote}
\usepackage{makecell}
\usepackage[capitalize,noabbrev]{cleveref}
\theoremstyle{plain}

\theoremstyle{definition}

\theoremstyle{remark}

\usepackage{enumitem}

\newcommand{\staple}{\textsc{STAPLE}\xspace}
\newcommand{\stem}{\textsc{STEM}\xspace}
\newcommand{\olmosmall}{OLMo-2 1B\xspace}
\newcommand{\olmomid}{OLMo-2 7B\xspace}
\newcommand{\dolma}{\textsc{Dolma 3 Dolmino Mix}\xspace}
\newcommand{\dclm}{\textsc{DCLM Baseline 1.0}\xspace}

\definecolor{lavenderpink}{rgb}{0.95, 0.85, 0.95}
\definecolor{softlavender}{rgb}{0.64, 0.50, 0.68}
\title{STAPLE: Static Sparse Upcycling of Pretrained Checkpoints via Embedding Modules}

\usepackage{tikz}
\usetikzlibrary{arrows.meta,positioning,fit,calc,shapes.geometric}

\tikzset{
  box/.style={rounded corners=6pt, draw, thick, inner sep=4pt, fill=#1},
  op/.style={box=#1, minimum width=2.1cm, minimum height=0.8cm},
  tinyop/.style={box=#1, minimum width=1.8cm, minimum height=.65cm},
  io/.style={box=#1, minimum width=1.4cm, minimum height=.65cm},
  circ/.style={circle, draw, fill=white, thick, minimum size=6mm},
  note/.style={draw=gray, dashed, rounded corners=4pt, inner sep=3pt},
  flow/.style={-{Latex[length=1.2mm,width=0.9mm]}, semithick, shorten >=1pt},
  dashedflow/.style={densely dashed, semithick, -{Latex[length=1.2mm,width=0.9mm]}},
  module/.style={rounded corners=10pt, draw=black!65, blur shadow,
                 inner sep=8pt, fill=white},
  subop/.style={rounded corners=8pt, draw=black!55, thick, inner sep=6pt, fill=#1},
  dottedflow/.style={densely dotted, -Latex, semithick},
  pill/.style={rounded corners=10pt, draw=black!60, thick, fill=#1, inner sep=6pt},
  op/.style={rounded corners=8pt, draw=black!65, thick, fill=#1, inner sep=7pt, minimum height=12mm},
  circop/.style={circle, draw=black!70, thick, fill=white, minimum size=7mm},
  panel/.style={rounded corners=10pt, draw=#1!65!black, thick, fill=#1!5,
                align=left, inner sep=8pt, text width=0.34\textwidth},
  tag/.style={rounded corners=6pt, draw=black!40, thin, fill=black!5, inner sep=3pt, font=\scriptsize\sffamily},
}

\author[\dagger]{Ranajoy Sadhukhan}
\author[\dagger]{Euijin Hong}
\author[\mathparagraph]{Sheng Cao}
\author[\mathsection]{Zechun Liu}
\author[\ddagger]{Shuowei Jin}
\author[\ddagger]{Sha Li}
\author[\ddagger]{Changlong Yu}
\author[\ddagger]{Xin Liu}
\author[\ddagger]{Qin Lu}
\author[\ddagger]{Zhenyu Shi}
\author[\ddagger]{Priyanka Nigam}
\author[\dagger]{Beidi Chen}

\affiliation[\dagger]{Carnegie Mellon University}
\affiliation[\mathparagraph]{D. E. Shaw Research}
\affiliation[\mathsection]{Meta AI}
\affiliation[\ddagger]{Amazon Research}

\contact{\email{rsadhukh@andrew.cmu.edu}}
\contact{\email{ejhong@andrew.cmu.edu}}
\contact{\email{rick.caos@gmail.com}}
\contact{\email{zechunliu@meta.com}}
\contact{\texttt{\{\href{mailto:shuowei@amazon.com}{shuowei},\allowbreak \href{mailto:slliz@amazon.com}{slliz},\allowbreak \href{mailto:changlyu@amazon.com}{changlyu},\allowbreak \href{mailto:xliucr@amazon.com}{xliucr},\allowbreak \href{mailto:luqn@amazon.com}{luqn},\allowbreak \href{mailto:zshi@amazon.com}{zshi},\allowbreak \href{mailto:nigamp@amazon.com}{nigamp}\}@amazon.com}}
\contact{\email{beidic@andrew.cmu.edu}}

\abstract{Sparse computation can scale the parametric capacity of large language models without proportionally increasing active computation, but pretraining sparse architectures remains expensive and data-hungry. Existing sparse upcycling methods mainly target Mixture-of-Experts (MoE) models, where expert homogeneity, sample inefficiency, and system overheads can limit practical scalability. Static sparse memory offers predictable access and token-specific updates that can help address these challenges, but naively integrating it into pretrained computation yields poor returns. In this work, we introduce \staple, a recipe built upon \stem for integrating static sparse memory into pretrained models. Whereas \stem replaces FFN up-projections with token-indexed embeddings, \staple retains these projections and adds the embeddings as a residual pathway, preserving pretrained computation while learning complementary token-specific information in a scalable manner. On \olmosmall, \staple improves over continued dense training by 4.27 points on HumanEval, 2.80 on MBPP, 2.27 on GSM8K, and 1.65 on DROP, with gains spanning programming, mathematical reasoning, and reading comprehension. Average performance increases with static capacity, and improvements persist on the larger \olmomid backbone. \staple also outperforms static sparse baselines such as \stem and an Engram-style model, and achieves a higher average score than an MoE upcycling baseline while decoding 2$\times$ faster with less than half its GPU model memory and measured GPU energy per token in our throughput-oriented decoding evaluation. Importantly, despite introducing billions of parameters late in training, our host-offloaded implementation keeps generation throughput within 1.5\% and training throughput within 3\% of the dense baseline, while caching only the embeddings of the $\sim$5\% most frequent tokens on the GPU. These results show that static sparse memory can provide a sample-efficient, scalable, and sustainable path for expanding pretrained models without sparse pretraining from scratch.
}
\metadata[Github]{\url{https://github.com/Infini-AI-Lab/STAPLE}}
\metadata[Website]{\url{https://infini-ai-lab.github.io/STAPLE}}

\begin{document}

\maketitle
\section{Introduction}

Scaling model capacity remains one of the most reliable ways to improve language-model capabilities~\citep{kaplan2020scalinglawsneurallanguage,hoffmann2022trainingcomputeoptimallargelanguage}. Sparse computation makes this scaling attractive by increasing total parametric capacity without proportionally increasing active computation. Yet repeatedly pretraining larger sparse models from scratch is prohibitively expensive. This motivates \emph{sparse upcycling}~\citep{komatsuzaki2023sparseupcyclingtrainingmixtureofexperts,he2025upcyclinglargelanguagemodels,vavre2024llama3meetsmoe}: expanding an already pretrained checkpoint through limited continued training such that the full potential of large parametric capacity is realized without training from scratch.


Most existing sparse upcycling methods rely on Mixture-of-Experts (MoE). A standard strategy clones pretrained FFNs into multiple experts, preserving the seed model at initialization~\citep{komatsuzaki2023sparseupcyclingtrainingmixtureofexperts,he2025upcyclinglargelanguagemodels}. However, cloned experts can remain functionally similar, limiting the benefit of further expert scaling. Breaking this \emph{expert homogeneity} through domain-specialized branches, partial reinitialization, or additional optimization stages can improve specialization but at the cost of larger training budget~\citep{nakamura2025dropupcyclingtrainingsparsemixture} or additional training stages~\citep{sukhbaatar2024branchtrainmixmixingexpertllms,mirvakhabova2025dirichletpriorshapingguidingexpert}. Furthermore, MoE scaling also incurs routing, load-balancing, expert-fragmentation, memory, and communication overheads that can hurt practical training and serving efficiency~\citep{doubov2024sparseupcyclinginference}. The value of added expert capacity therefore depends on its performance gains relative to these costs. Hence, an effective upcycling method should be \textbf{sample-efficient}, preserve the \textbf{training and serving efficiency} of the seed model, and remain \textbf{scalable} as sparse capacity grows.

\begin{figure*}[t]
    \centering
    \subfloat[MMLU score vs train tokens]{
        \includegraphics[width=0.31\linewidth]{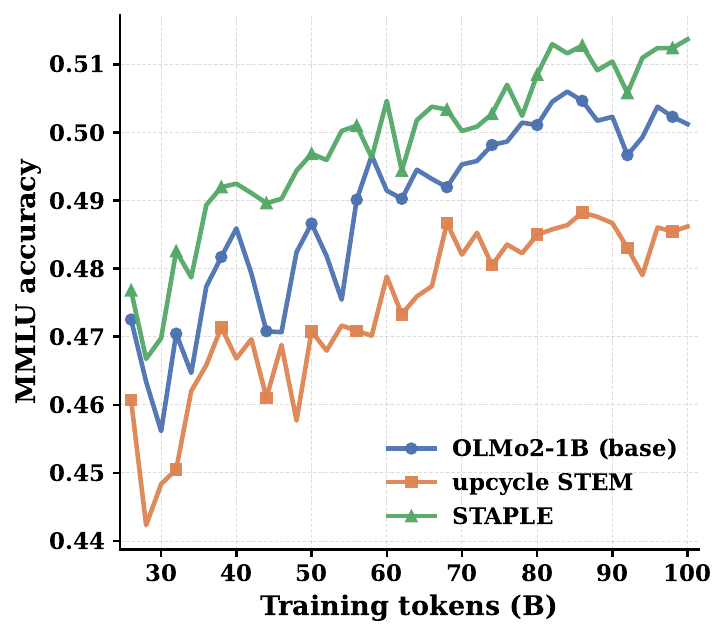}
        \label{fig:token-efficiency}
    }
    \subfloat[Scalability]{
        \includegraphics[width=0.31\linewidth]{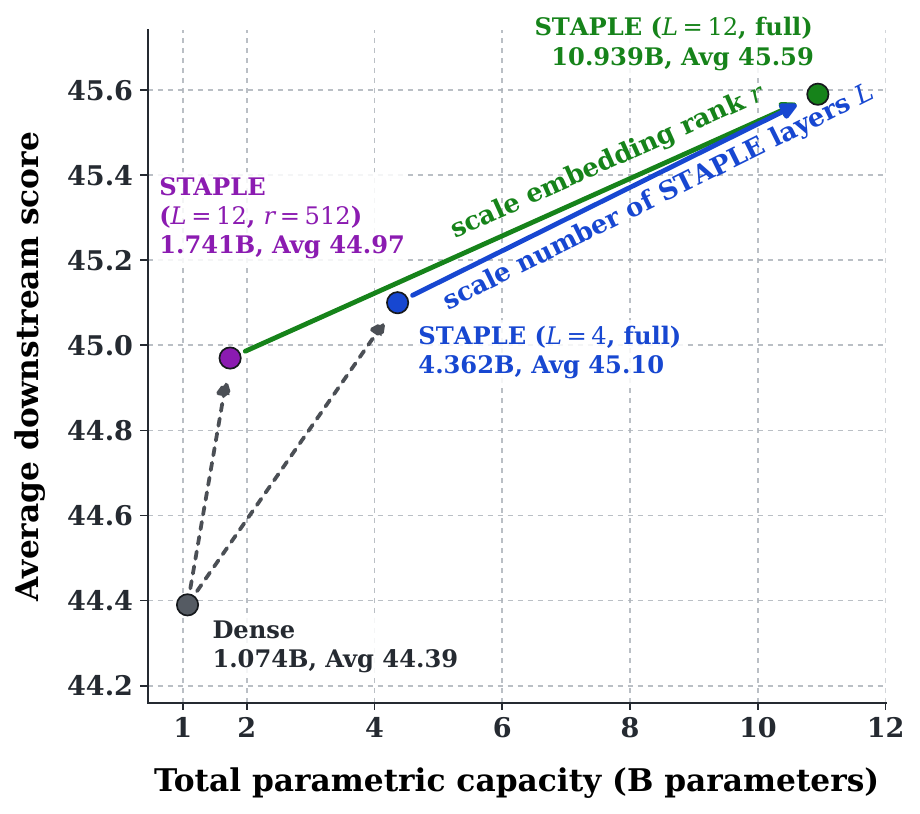}
        \label{fig:scalability}
    }
    \subfloat[Comparative Analysis]{
        \includegraphics[width=0.36\linewidth]{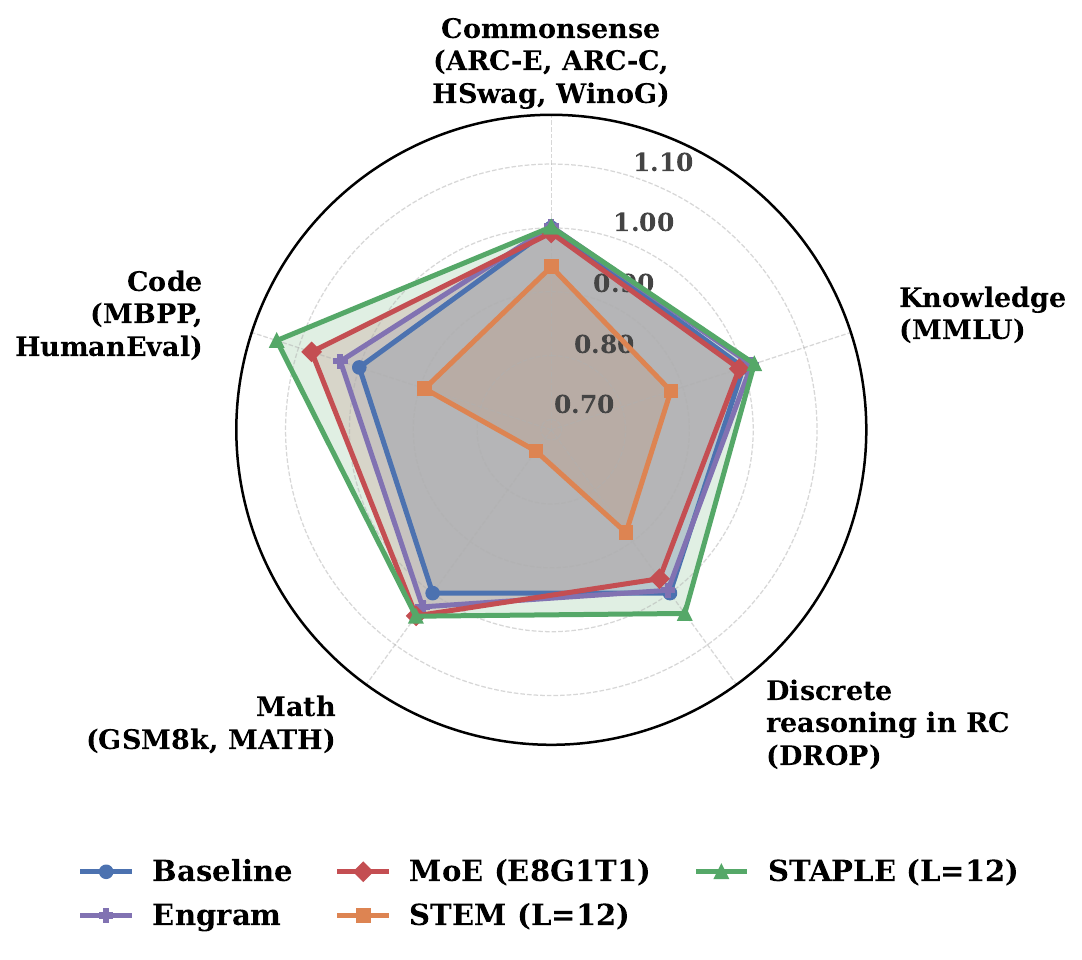}
        \label{fig:overall-perf}
    }
    \caption{\textbf{Overview of \staple.} \textbf{(a)} \staple improves over the dense baseline early in mid-training, while replacement-style \stem upcycling remains below it. The horizontal axis counts mid-training tokens, excluding the preceding warmup or dense continuation stage. \textbf{(b)} Downstream performance improves as static sparse capacity is increased through embedding rank or the number of augmented layers. \textbf{(c)} Domain averages normalized to continued dense training; the largest gain is in code, with smaller gains in mathematics, knowledge, and reading comprehension.}
\end{figure*}

Recent emergence in static sparse architectures~\citep{gemma3n2024,sadhukhan2026stemscalingtransformersembedding,cheng2026conditionalmemoryscalablelookup,liu2026scalingembeddingsoutperformsscaling,tseng2026l3largelookuplayers} has brought a breakthrough in scaling parametric capacity in a system-efficient manner. These architectures insert large token- or phrase-indexed embedding tables that are particularly amenable to CPU offloading and asynchronous prefetching, allowing substantial parametric expansion with little additional GPU-memory footprint or latency overhead. Unlike cloned MoE experts, token-indexed rows receive updates from distinct token-conditioned subsets of the training data, providing a structural bias toward differentiation.
However, static sparse upcycling introduces its own challenges. \textit{First}, replacing pretrained computation with newly initialized sparse memory discards useful features that must then be relearned. \textit{Second}, existing static sparse architectures can suffer from poor sample efficiency: each embedding row is updated only when its corresponding token or phrase appears, producing highly \textit{non-uniform training signals} under a limited token budget. \textit{Finally}, even if these optimization challenges are addressed, it remains unclear whether static sparsity can continue to deliver performance gains as additional sparse capacity is introduced. Our key findings below directly address these concerns.

\underline{\textit{Sparsity should complement existing architecture:}}
Retaining pretrained computation is important for effective static sparse upcycling, which is often not maintained by existing static sparse approaches. For instance, STEM~\citep{sadhukhan2026stemscalingtransformersembedding} replaces the FFN up-projection with a token-indexed embedding table. As Fig.~\ref{fig:token-efficiency} illustrates, \stem upcycling continues to struggle with downstream performance even after 100B tokens, whereas retaining the pretrained FFN up-projection and adding static memory as a residual pathway performs substantially better. This allows the new parameters to focus on complementary token-specific information rather than relearning computation already present in the seed model.

\begin{figure*}[t]
    \centering
    \subfloat[Common sense reasoning ablation]{
        \includegraphics[width=0.32\linewidth]{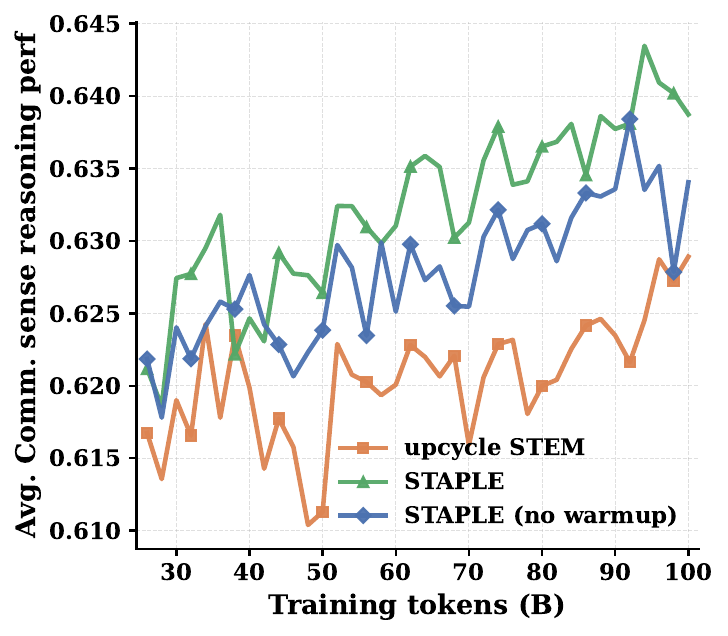}
        \label{fig:ablation_comm_sense}
    }
    \subfloat[Math \& code ablation]{
        \includegraphics[width=0.32\linewidth]{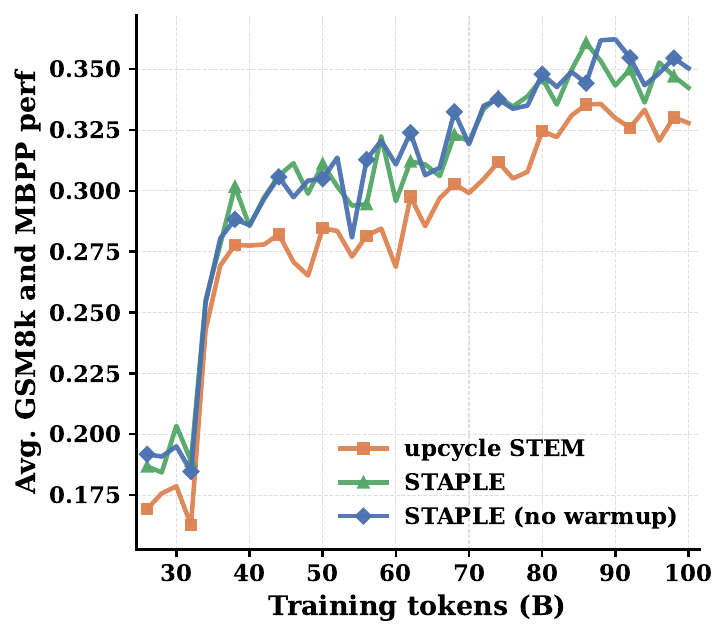}
        \label{fig:ablation_math_code}
    }
    \subfloat[MMLU ablation]{
        \includegraphics[width=0.32\linewidth]{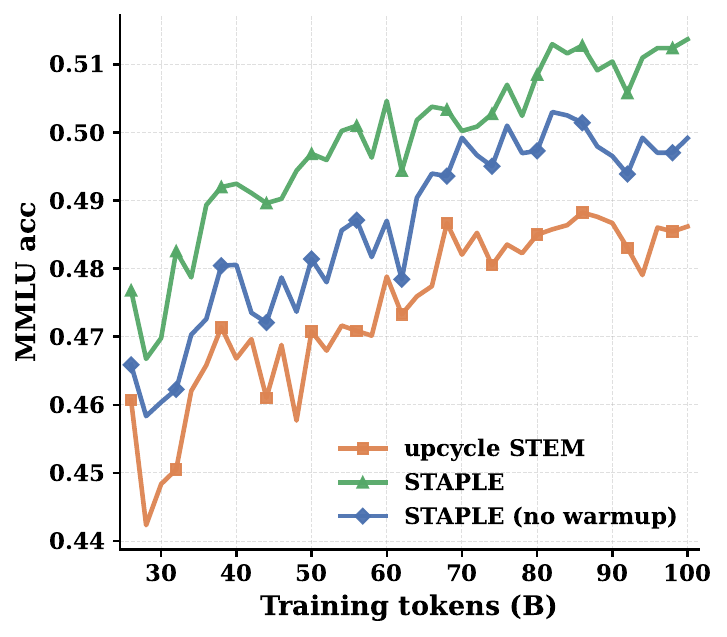}
        \label{fig:ablation_mmlu}
    }
    \caption{Domain-specific performance comparison among \stem upcycling, \staple and \staple with no warmup. Each model uses four embedding layers. The horizontal axes count mid-training tokens, excluding the separate warmup stage where used. Panels (a--c) illustrate the importance of retaining the FFN up-projections in \staple as it consistently outperforms \stem in every aspect. Panels (a,c) show the importance of warmup in improving general knowledge and commonsense task performances. Conversely, panel (b) suggests that math and code reasoning performances are relatively less affected.}
    \label{fig:ablation}
\end{figure*}

\underline{\textit{Initialize for sample-efficient adaptation:}}
Preserving the pretrained model's initial functionality through correct initialization is important for better sample efficiency.  Additionally, a short warmup with selective freezing of pretrained components then encourages the newly introduced embeddings to learn useful complementary knowledge, particularly improving knowledge-intensive and commonsense tasks (Figs.~\ref{fig:ablation_mmlu} and~\ref{fig:ablation_comm_sense}).

\underline{\textit{Scalability:}}
When allowed to learn complementary token-specific knowledge, static memory continues to improve performance as capacity grows without proportionally increasing systems cost (Fig.~\ref{fig:sys_efficiency}). As illustrated in Fig.~\ref{fig:scalability}, \staple improves in average downstream performance as we increase either the number of augmented layers or the embedding dimension, while the added memory can remain largely offloaded to CPU memory. 


Based on these observations, we introduce \textbf{STAtic uPcycLing via Embedding layers (\staple)}. Building on \stem~\citep{sadhukhan2026stemscalingtransformersembedding}, \staple retains the pretrained FFN up-projection and augments it with a zero-initialized token-indexed embedding. The dense pathway preserves shared pretrained computation, while the static pathway learns complementary token-specific information. We scale this capacity through both the number of augmented layers and the embedding dimension. For systems efficiency, the large embedding tables remain in CPU memory and are accessed through asynchronous prefetching, LFU caching, token deduplication, and row-sparse optimization during training and serving.

On \olmosmall, \staple improves over continued dense training by 4.27 points on HumanEval, 2.80 on MBPP, 2.27 on GSM8K, and 1.65 on DROP, raising the ten-task average by 1.20 points (Fig.~\ref{fig:overall-perf}). Average performance improves with static capacity (Fig.~\ref{fig:scalability}), and gains persist on the larger \olmomid backbone. Under the same budget, \staple also outperforms the parameter-matched static sparse baselines, \stem and an Engram-style model. Compared with a serving-efficient MoE upcycling baseline, \staple achieves a higher average score while decoding 2$\times$ faster with a much smaller GPU memory footprint, and it stays within 1.5\% and 3\% of dense generation and training throughput. Furthermore, predictable memory access could further benefit speculative decoding~\citep{leviathan2023fastinferencetransformersspeculative} by enabling prefetching of the rows needed for target verification~\citep{liu2026scalingembeddingsoutperformsscaling}. The reduced HBM footprint leaves more room for KV caching and larger batches, while differences in weight reuse and execution overhead motivate examining energy alongside parameter count: in our throughput-oriented decoding evaluation, \staple keeps measured GPU energy per token within 1--2\% of dense and reduces it by over 50\% relative to the MoE baseline. Beyond these performance and efficiency gains, \staple exposes an interpretable and editable memory: targeted embedding interventions modify model behavior while preserving its ability to follow contextual evidence (Sec.~\ref{subsec:interpretability}). Together, these results suggest that existing pretrained models can gain new capabilities from \textit{scalable} static sparse memory without expensive training or a prohibitive increase in serving cost.

\section{Method}
Our final method builds upon three key aspects. Sec.~\ref{sec: arch study} studies the importance of sparse architecture selection. Sec.~\ref{sec:initialization} discusses what initialization of the embedding tables can restore the seed model's performance and what its implications are in the downstream task performance. Finally, Sec.~\ref{sec:scaling} illustrates different scaling axes of \staple and their scaling behavior.

\subsection{Architecture study: impact of retaining the up-projection}
\label{sec: arch study}

\paragraph{STEM Layers.}
Consider a gated FFN with gate, up, and down projections $W^g_\ell$, $W^u_\ell$, and $W^d_\ell$, respectively, and hidden state $x_\ell$ at layer $\ell$. STEM~\citep{sadhukhan2026stemscalingtransformersembedding} replaces the pretrained up-projection output $W^u_\ell x_\ell$ with a token-indexed embedding $U_\ell[t]$, where $t$ is the input token id and $U_\ell\in\mathbb{R}^{V\times d_{\mathrm{ff}}}$ is a layer-local table with vocabulary size $V$ and FFN width $d_{\mathrm{ff}}$:
\begin{equation}
    y^{\mathrm{STEM}}_\ell
    =
    W^d_\ell\left(\mathrm{SiLU}(W^g_\ell x_\ell)\odot U_\ell[t]\right).
    \label{eq:stem_replacement}
\end{equation}

A key architectural choice is whether sparse memory should replace or augment pretrained computation. In our four-layer ablation, replacement-based \stem upcycling fails to match continued dense training even with additional warmup (Fig.~\ref{fig:token-efficiency}), whereas retaining the pretrained FFN up-projection and adding \stem embeddings as a residual pathway yields substantially better downstream performance under the same training budget (Fig.~\ref{fig:ablation}).

Token-indexed embeddings receive updates only when their corresponding tokens appear, producing highly non-uniform training signals under a limited token budget. Replacement-based \stem compounds this general challenge of sparse memory by discarding useful pretrained up-projection features, which the new embeddings must then relearn. Moreover, it lacks a straightforward functionality-preserving initialization: random, token-wise mean, principal-component, and transformed input-embedding initializations did not improve long-term performance in our experiments.

\begin{figure*}[t]
  \centering
    \includegraphics[width=\linewidth,keepaspectratio]{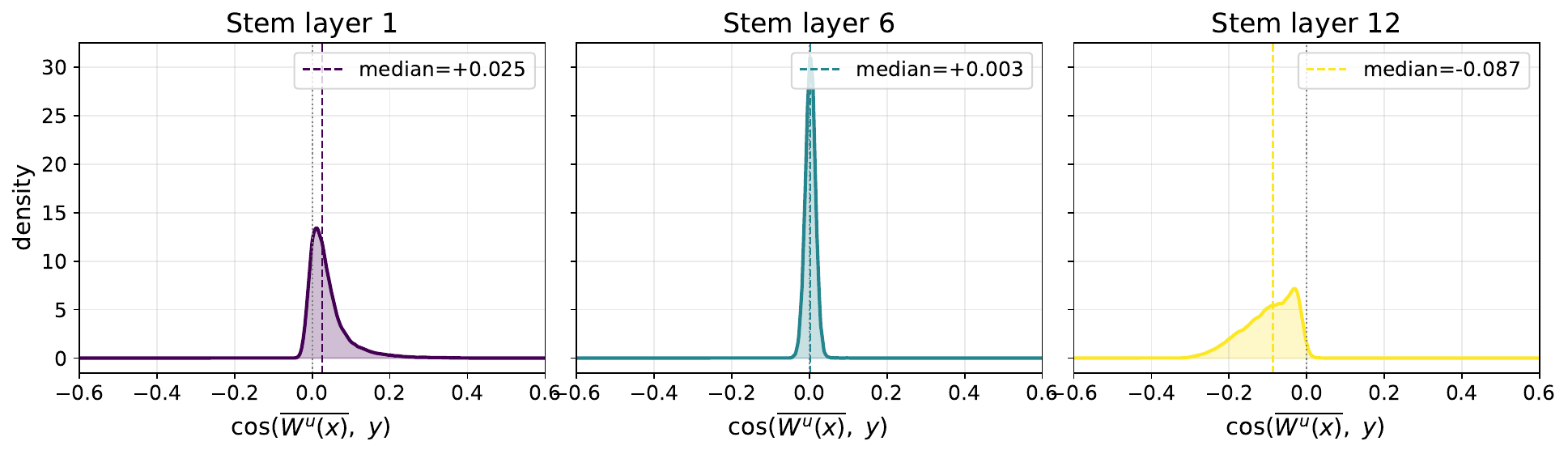}

  \caption{Pair-wise cosine similarity analysis between mean token-wise up-projection output and \stem embeddings in different layers of \staple with 12 \stem tables. Overall, the \stem embeddings are quite orthogonal to the respective mean up-projection output vectors.}
  \label{fig:complementary}
\end{figure*}

These observations motivate the additive design of \staple:
\begin{equation}
    y^{\mathrm{STAPLE}}_\ell
    =
    W^d_\ell
    \left(
        \mathrm{SiLU}(W^g_\ell x_\ell)
        \odot
        \left(W^u_\ell x_\ell + U_\ell[t]\right)
    \right).
    \label{eq:staple_additive}
\end{equation}
The pretrained up-projection preserves shared dense computation, while the added embeddings learn complementary token-specific residual information, as further examined in Sec.~\ref{sec:analysis_learned_embeddings} and Fig.~\ref{fig:complementary}.

\subsection{Initialization}
\label{sec:initialization}

We initialize the embeddings with zero weights. This restores the seed model's computation graph and avoids discarding the knowledge embedded into the pretrained model weights. 

\paragraph{Warmup.}
Furthermore, we find that a short warmup phase can better prepare the newly introduced \stem embeddings for subsequent mid-training. Before training on the high-quality mid-training mixture, we train \staple on a small subset of the pretraining corpus ($\approx 2.5\%$). Since the seed model has already been optimized on this corpus, we freeze the FFN up-projections during warmup. This preserves the pretrained dense computation while forcing the new embedding tables to learn useful complementary signals.

As shown in Fig.~\ref{fig:ablation_mmlu} and Fig.~\ref{fig:ablation_comm_sense}, warmup consistently improves MMLU and average commonsense reasoning over direct upcycling throughout mid-training. This suggests that brief exposure to the pretraining distribution helps token-indexed embeddings acquire useful general-purpose memory before adapting to the mid-training mixture. In contrast, warmup provides no consistent gains on advanced math and programming tasks (Fig.~\ref{fig:ablation_math_code}). To control this comparison, we exclude math and code sources from warmup, although they constitute a small fraction of the original pretraining corpus. These results suggest that sparse memory remains responsive to the subsequent mid-training distribution and can learn capabilities not emphasized during warmup.

\subsection{Scalability}
\label{sec:scaling}
We scale \staple along two axes: the number of embedding-augmented layers $L$ and the embedding dimension $r$. Under the same continued-training budget, increasing $L$ from 4 to 12 raises the average downstream score from 45.10 to 45.59 (Table~\ref{tab:staple_scalability}; Fig.~\ref{fig:scalability}). The gains include reasoning-heavy tasks: from $L=4$ to $L=12$, MBPP improves by 1.80 points, HumanEval by 1.83 points, and DROP by 0.73 points, while MMLU is essentially unchanged (51.27 vs.\ 51.21). We place the embedding tables in early decoder layers (1--12), allowing token-specific residuals to be integrated into representations processed by subsequent layers.

To vary the memory capacity within each layer, we replace the full-rank table $U_\ell\in\mathbb{R}^{V\times d_{\mathrm{ff}}}$ with $U^{(r)}_\ell\in\mathbb{R}^{V\times r}$ and learn a projection $P_\ell\in\mathbb{R}^{d_{\mathrm{ff}}\times r}$:
\begin{equation}
    y^{\mathrm{STAPLE}(r)}_\ell
    =
    W^d_\ell
    \left(
        \mathrm{SiLU}(W^g_\ell x_\ell)
        \odot
        \left(
            W^u_\ell x_\ell + P_\ell U^{(r)}_\ell[t]
        \right)
    \right).
    \label{eq:staple_low_rank}
\end{equation}
The resulting module has $Vr+d_{\mathrm{ff}}r$ parameters per layer and adds $2d_{\mathrm{ff}}r$ projection FLOPs per token. For \olmosmall, $r=512$ reduces the token-indexed table by $16\times$. At $L=12$, this low-rank variant reaches 44.97 average performance versus 45.59 for full-rank \staple and 44.39 for dense continuation (Table~\ref{tab:staple_scalability}). Together, these results show that \staple benefits from scaling both the number of augmented layers and the capacity of each embedding table.

\subsection{System Efficiency}
\label{sec: sys efficiency}
Although \staple substantially increases the parametric memory footprint, the predictable token-indexed access pattern allows the large tables to remain in CPU DRAM, with only needed rows transferred to the GPU for training and inference. The challenge is to hide the host to device memory transfer cost behind model computation.

\begin{figure*}[t]
    \centering
    \subfloat[Training throughput and GPU memory]{%
        \includegraphics[width=0.64\linewidth]{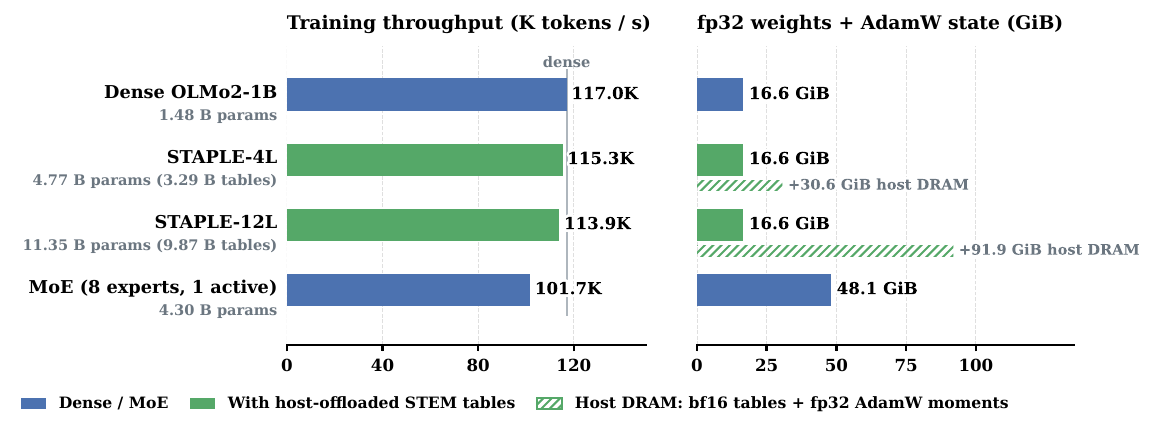}%
        \label{fig:train_throughput}%
    }\hfill
    \subfloat[Decode throughput]{%
        \includegraphics[width=0.34\linewidth]{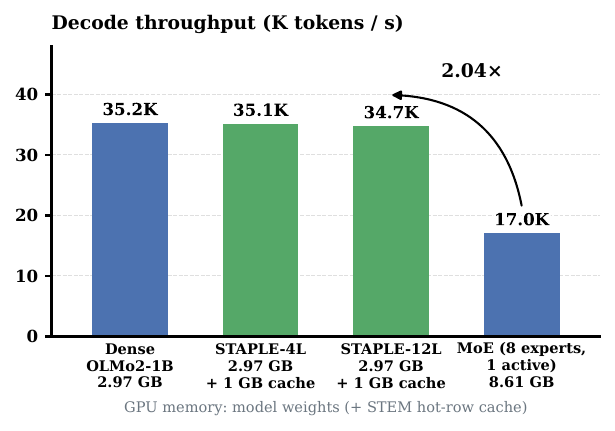}%
        \label{fig:decode_throughput}%
    }
    \caption{\textbf{Training and serving efficiency of host-offloaded \staple.} \textbf{(a)} Training throughput and GPU-resident weights plus optimizer state; parameter counts include the \stem tables. \staple keeps its bf16 tables and fp32 AdamW moments in host DRAM (hatched), matching the dense GPU footprint, whereas the MoE baseline (Sec.~\ref{sec:setup}) keeps all experts and their optimizer states on the GPU. \textbf{(b)} Decode throughput on one NVIDIA B200 at batch size 64 and context length 256, labeled with GPU memory for weights plus, for \staple, a 1\,GB cache of frequently accessed rows. Latency and prefill results are in Table~\ref{tab: inference table}.}
    \label{fig:sys_efficiency}
\end{figure*}

\paragraph{Inference-time offloading.}
We reduce retrieval overhead through three complementary optimizations. \textbf{Asynchronous prefetching} overlaps transfers for upcoming requests or prefill chunks with current computation; during decode, later-layer lookups overlap with earlier-layer execution. \textbf{Frequency-aware caching} exploits the Zipfian token distribution. Caching 5--6\% of the vocabulary achieves roughly 80\% hit rate, reducing the GPU-resident \staple-12L embedding footprint from 19.73\,GB to a 1\,GB cache. \textbf{Token deduplication} transfers each distinct row only once per prefill batch, reducing prefetch workload by roughly 50\%. We omit it during decode, where the deduplication overhead exceeds the savings.
As Fig.~\ref{fig:decode_throughput} shows, with 19.73\,GB of learned tables offloaded, \staple-12L stays within 1.5\% of dense decode throughput at batch size 64 (within 1.0\% at batch size 1; Table~\ref{tab: inference table}). Prefill throughput is within 1.5\% of dense at batch size 64, but is 6.9\% lower at batch size 1. Relative to the MoE baseline, \staple uses about 54\% less HBM model memory and improves decode throughput by 2.04$\times$ at batch size 64 and 58\% at batch size 1. Beyond latency, energy per generated token captures the benefits of weight reuse across the batch, efficient dense kernels, and reduced background energy from shorter execution. In our throughput-oriented decoding evaluation, \staple's measured GPU energy per token stays within 1--2\% of dense and over 50\% below the evaluated MoE implementation (Appendix~\ref{app:gpu_energy}). The same predictable access pattern could also benefit speculative decoding, as draft token IDs expose the rows needed for target verification, allowing embedding prefetching to overlap with ongoing computation.

\paragraph{Training-time offloading.}
\label{sec: offload}
Training also requires gradients and optimizer states, making full-table GPU residency costly. Yet each step accesses only the \textbf{distinct token ids} in its batch, $U=|\mathrm{uniq}(\mathcal{B})|$, rather than the full vocabulary $V$. With non-uniform token frequencies, typically $U\ll V$, and $U$ grows sublinearly with batch size~\citep{takahashi2019evaluatingcomputationallanguagemodels}. We keep the full \stem tables and fp32 AdamW states in CPU DRAM and materialize only the active $U\times d_{\mathrm{ff}}$ rows on GPU. This compact buffer acts as the autograd leaf, avoiding dense $V\times d_{\mathrm{ff}}$ gradients; row-restricted AdamW updates only the accessed rows. We observe no degradation in end-to-end downstream task performance when training with these sparse optimizer updates. To improve throughput, we overlap host gathering, host-to-device/device-to-host transfers, and optimizer work with backbone computation. As host work scales with $U$ but transformer compute scales with the total token count, overlap becomes more favorable at larger batches. \staple-12L trains within 3\% of dense throughput (113.9K vs.\ 117.0K tokens/s) with 91.9\,GiB of tables and optimizer moments in host DRAM; its GPU-resident weights and optimizer state match the dense model, and the compact row buffers add only about 3\% peak HBM, whereas the MoE baseline trains 13\% slower (101.7K tokens/s) and needs 2.9$\times$ the GPU memory for weights and optimizer state (Fig.~\ref{fig:train_throughput}). Appendix~\ref{sec: implementation} details both pipelines.
\section{Experimental Setup}
\label{sec:setup}
\paragraph{Models.}
We use \olmosmall~\citep{olmo20242olmo2furious} as the primary dense seed model because its training recipe, datasets, and optimizer-state checkpoints are publicly available. The released checkpoint is pretrained on 4T tokens and contains 16 decoder layers. We construct \staple variants by augmenting the early decoder layers with 4 or 12 \stem embedding tables, all zero-initialized so that the upcycled model initially preserves the computation and performance of the dense checkpoint. Additionally, we upcycle the same \olmosmall backbone with a single-stream Engram-style model using memory modules at layers 2 and 7, hashed 2-gram and 3-gram memories with eight 64-dimensional heads per order, and approximately 9.87B additional embedding-memory parameters matched to \staple-12L; this implementation uses context-conditioned key/value projections with sigmoid gating and a depthwise causal convolution, but omits Engram's multi-branch hyper-connections. Following the upcycling approach of \citet{he2025upcyclinglargelanguagemodels}, we also include a relatively serving-efficient MoE baseline, which replaces the FFNs in eight interleaved decoder layers with MoE layers of eight full-width experts and top-1 routing; we refer to it as the MoE baseline.

\paragraph{Datasets.}
For mid-training, we use the \dolma~\citep{olmo2025olmo3} high-quality data mixture, which contains training sequences from web, non-web, and synthetic sources. This mixture is designed to improve academic question answering, mathematical reasoning, and programming abilities, and to prepare models for subsequent post-training~\citep{ouyang2022traininglanguagemodelsfollow, wei2022finetunedlanguagemodelszeroshot}. For the additional warmup phase, we train on approximately 100B tokens sampled from \dclm~\citep{li2025datacomplmsearchgenerationtraining}, the main component of the seed model's pretraining mixture. This budget is approximately 2.5\% of the seed checkpoint's 4T pretraining tokens. The original mixture also includes dedicated code and math sources~\citep{olmo20242olmo2furious}, which we exclude from warmup. This choice allows us to isolate the effect of warmup on \stem initialization without introducing additional math or code-specific supervision before mid-training.

\paragraph{Training.}
For all dense backbone parameters, we follow the \olmosmall training hyperparameters~\citep{olmo20242olmo2furious} and initialize the optimizer from the released pretrained optimizer states. During mid-training, newly introduced \stem embeddings use a peak learning rate approximately 4$\times$ that of the dense parameters, and all parameter groups follow a linear decay to zero. Warmup instead uses cosine schedules, with the backbone resuming its original pretraining schedule. Full training hyperparameters are listed in Table~\ref{tab:training-hyperparams} (Appendix~\ref{app:hyperparams}). During the \staple warmup phase, we freeze the up-projections in the augmented layers so that the newly added static memory can adapt without perturbing this pretrained pathway. When the separate \stem warmup stage is omitted, we instead apply a 2\% warmup ratio to the embedding-specific learning-rate schedule during mid-training.
The main 1B comparisons use approximately 100B warmup tokens followed by 100B mid-training tokens. The dense baseline receives the same token budget, using its original pretraining learning-rate schedule during the warmup-equivalent stage. The 7B experiments use a reduced budget of approximately 30B warmup tokens followed by 70B mid-training tokens due to their higher computational cost. 

\paragraph{Evaluations.}
We evaluate our models on a broad benchmark suite spanning commonsense reasoning, general knowledge, reading comprehension, mathematics, and coding. For commonsense reasoning, we report ARC-Easy~\citep{allenai:arc} (0-shot, accuracy), ARC-Challenge~\citep{allenai:arc} (0-shot, normalized accuracy), HellaSwag~\citep{zellers2019hellaswag} (0-shot, normalized accuracy), and WinoGrande~\citep{ai2:winogrande} (0-shot, accuracy). We evaluate general knowledge on MMLU~\citep{hendryckstest2021} (0-shot, accuracy) and reading comprehension on DROP~\citep{Dua2019DROP} (3-shot, F1). For mathematics, we use GSM8K~\citep{cobbe2021gsm8k} (5-shot, strict exact match) and Hendrycks MATH~\citep{hendrycks2021measuringmathematicalproblemsolving} (0-shot, exact match). For coding, we evaluate MBPP~\citep{austin2021program} (5-shot, pass@1) and HumanEval~\citep{chen2021evaluating} (0-shot, pass@1).

\section{Results \& Analysis}
\begin{table*}[t]
\centering
\caption{\textbf{Upcycling comparison and design ablations.} All models in the first block use the \olmosmall seed; the Engram-style baseline's added embedding memory is parameter-matched to \staple-12L. The replacement-style \stem and additive \staple models both use 12 embedding layers. The second block reports the \olmomid backbone with $L=16$ for \staple. Best values are bolded within each backbone block.}
\label{tab:main_results}
\scriptsize
\setlength{\tabcolsep}{3.5pt}
\renewcommand{\arraystretch}{1.08}
\resizebox{\textwidth}{!}{%
\begin{tabular}{@{}lccccccccccc@{}}
\toprule
& \multicolumn{4}{c}{Commonsense}
& \multicolumn{1}{c}{Knowledge}
& \multicolumn{1}{c}{RC}
& \multicolumn{2}{c}{Math}
& \multicolumn{2}{c}{Code}
& \\
\cmidrule(lr){2-5}
\cmidrule(lr){6-6}
\cmidrule(lr){7-7}
\cmidrule(lr){8-9}
\cmidrule(lr){10-11}
\textbf{Method}
& \textbf{ARC-E}
& \textbf{ARC-C}
& \textbf{HSwag}
& \textbf{WinoG}
& \textbf{MMLU}
& \textbf{DROP}
& \textbf{GSM8K}
& \textbf{MATH}
& \textbf{MBPP}
& \textbf{HEval}
& \textbf{Avg.} \\
\midrule
\multicolumn{12}{@{}l}{\textit{\olmosmall: baselines and architecture/initialization comparisons}} \\
\addlinespace[1pt]
Baseline
& 75.42 & \textbf{47.44} & 65.18 & \textbf{66.22} & 50.30 & 41.91 & 37.76 & 7.68 & 28.80 & 23.17 & 44.39 \\
MoE upcycling (8 experts, top-1)
& 75.04 & 45.73 & 66.82 & 64.72 & 49.96 & 40.74 & 39.80 & 7.64 & 28.00 & \textbf{28.05} & 44.65 \\
Engram-style
& \textbf{77.40} & 47.35 & 66.05 & 64.01 & 50.98 & 41.68 & 38.21 & \textbf{8.44} & 29.20 & 24.39 & 44.77 \\
\stem ($L{=}12$)
& 74.37 & 42.06 & 60.79 & 61.56 & 44.30 & 37.00 & 26.00 & 6.94 & 22.60 & 23.78 & 39.94 \\
\staple ($L{=}12$)
& 76.60 & 45.99 & \textbf{68.27} & 63.77 & \textbf{51.21} & \textbf{43.56} & \textbf{40.03} & 7.44 & \textbf{31.60} & 27.44 & \textbf{45.59} \\
\midrule
\multicolumn{12}{@{}l}{\textit{\olmomid: larger-backbone evaluation}} \\
\addlinespace[1pt]
Baseline
& 82.07 & 55.97 & 77.52 & \textbf{74.82} & 62.53 & 62.79 & 67.79 & 11.08 & 41.80 & 34.76 & 57.11 \\
\staple ($L{=}16$)
& \textbf{82.58} & \textbf{57.76} & \textbf{77.62} & 74.35 & \textbf{62.67} & \textbf{63.55} & \textbf{70.13} & \textbf{12.02} & \textbf{43.40} & \textbf{35.98} & \textbf{58.01} \\
\bottomrule
\end{tabular}%
}
\end{table*}
Table~\ref{tab:main_results} shows that \staple-12L raises the ten-task average from 44.39 to 45.59 over continued dense training. Although large parametric memory is usually associated with memorization, the gains extend to reasoning-heavy tasks: HumanEval improves by 4.27 points, MBPP by 2.80, GSM8K by 2.27, and DROP by 1.65.

Under the same upcycling budget, \stem struggles to even match continued dense training performance, with notable performance degradations in knowledge-based tasks as well as GSM8K. This establishes the benefits of retaining the pretrained model's computation graph while upcycling with static memory.
Furthermore, \staple achieves higher domain-average math and code scores and higher DROP performance than the parameter-matched Engram-style baseline, supporting the effectiveness of its architecture and adaptation recipe. Compared with the MoE baseline, \staple gains 0.94 points on average while delivering higher decode throughput and using less GPU-resident model memory (Fig.~\ref{fig:sys_efficiency}). These comparisons establish a favorable performance--efficiency tradeoff for the evaluated configurations.

\begin{table*}[t]
\centering
\caption{\textbf{\staple scalability on \olmosmall.} Increasing the number of augmented layers $L$ and the embedding dimension $r$ are separate capacity axes; $r=512$ denotes the low-rank variant, while the other \staple rows use full-width embeddings. All rows use the same continued-training budget. Best values are bolded within this table.}
\label{tab:staple_scalability}
\scriptsize
\setlength{\tabcolsep}{3.5pt}
\renewcommand{\arraystretch}{1.08}
\resizebox{\textwidth}{!}{%
\begin{tabular}{@{}lccccccccccc@{}}
\toprule
& \multicolumn{4}{c}{Commonsense}
& \multicolumn{1}{c}{Knowledge}
& \multicolumn{1}{c}{RC}
& \multicolumn{2}{c}{Math}
& \multicolumn{2}{c}{Code}
& \\
\cmidrule(lr){2-5}
\cmidrule(lr){6-6}
\cmidrule(lr){7-7}
\cmidrule(lr){8-9}
\cmidrule(lr){10-11}
\textbf{Method}
& \textbf{ARC-E}
& \textbf{ARC-C}
& \textbf{HSwag}
& \textbf{WinoG}
& \textbf{MMLU}
& \textbf{DROP}
& \textbf{GSM8K}
& \textbf{MATH}
& \textbf{MBPP}
& \textbf{HEval}
& \textbf{Avg.} \\
\midrule
Baseline
& 75.42 & \textbf{47.44} & 65.18 & \textbf{66.22} & 50.30 & 41.91 & 37.76 & \textbf{7.68} & 28.80 & 23.17 & 44.39 \\
\staple ($L{=}4$)
& \textbf{76.98} & 46.25 & 66.00 & 65.82 & \textbf{51.27} & 42.83 & 39.35 & 7.08 & 29.80 & 25.61 & 45.10 \\
\staple ($L{=}12$)
& 76.60 & 45.99 & \textbf{68.27} & 63.77 & 51.21 & \textbf{43.56} & \textbf{40.03} & 7.44 & \textbf{31.60} & \textbf{27.44} & \textbf{45.59} \\
\staple ($L{=}12$, $r{=}512$)
& 76.18 & 46.67 & 66.99 & 65.59 & 50.49 & 41.91 & 39.80 & 6.28 & 30.20 & 25.61 & 44.97 \\
\bottomrule
\end{tabular}%
}
\end{table*}

Table~\ref{tab:staple_scalability} separates the effects of augmenting more layers and increasing the embedding dimension. \staple improves average performance along both scaling axes, and the added capacity has negligible impact on model throughput.
We additionally evaluate \staple on an \olmomid backbone to test whether the gains persist beyond the 1B setting. We augment the first 16 decoder layers with \stem tables and, due to the substantially higher training cost, restrict training to approximately 30B warmup tokens followed by 70B mid-training tokens. As shown in Table~\ref{tab:main_results}, \staple improves the average score from 57.11 to 58.01 and improves 9 of the 10 evaluated tasks. The gains are particularly encouraging on reasoning and code: average mathematics performance improves by 4.16\% relative to the 7B baseline, while average code performance improves by 3.68\%. The modest MMLU gain may reflect the shorter warmup, consistent with the 1B warmup ablation, although this explanation is not isolated in the 7B experiments. These results suggest that the benefits of static sparse memory extend to larger backbones even under a substantially reduced training budget.

\subsection{Ablation studies}
\label{sec:analysis_learned_embeddings}

Figure~\ref{fig:complementary} compares learned token embeddings with their corresponding mean up-projection outputs. Their low cosine similarity suggests that the two feature streams occupy different directions, but geometry alone does not establish the memory's usefulness. We therefore test the model's dependence on these embeddings through removal ablations.

Specifically, we progressively remove the last 4, last 8, or all 12 embedding tables from a \staple-12L checkpoint at evaluation time, without retraining. As shown in Fig.~\ref{fig:layer_ablation}, the average score across ten benchmarks falls from 45.59 to 41.28, 36.94, and 32.17, respectively. The sharp decline shows that the model relies on the learned static memory after mid-training. The fully ablated model is not equivalent to the original dense seed, since the remaining weights have co-adapted with static memory.


\paragraph{Dependence on embedding depth.} The dependence is especially pronounced on math and code tasks: removing all tables lowers GSM8K, MBPP, and HumanEval by 31.08, 27.40, and 23.17 points, compared with 7.34 on ARC-Challenge and 7.08 on MMLU. Even removing only the last four tables reduces GSM8K by 17.59 points, MBPP by 8.40 points, and HumanEval by 8.54 points. Thus, the learned memory contributes beyond the earliest augmented layers, particularly to generation-heavy mathematical and programming tasks, whereas the knowledge-intensive tasks benefit more from the bottom-most embedding layers.

\subsection{Interpretability: Interplay between static memory and contextual reasoning}
\label{subsec:interpretability}

\begin{figure*}[t]
    \centering
    \subfloat[Embedding layer ablation]{
        \includegraphics[width=0.32\linewidth]{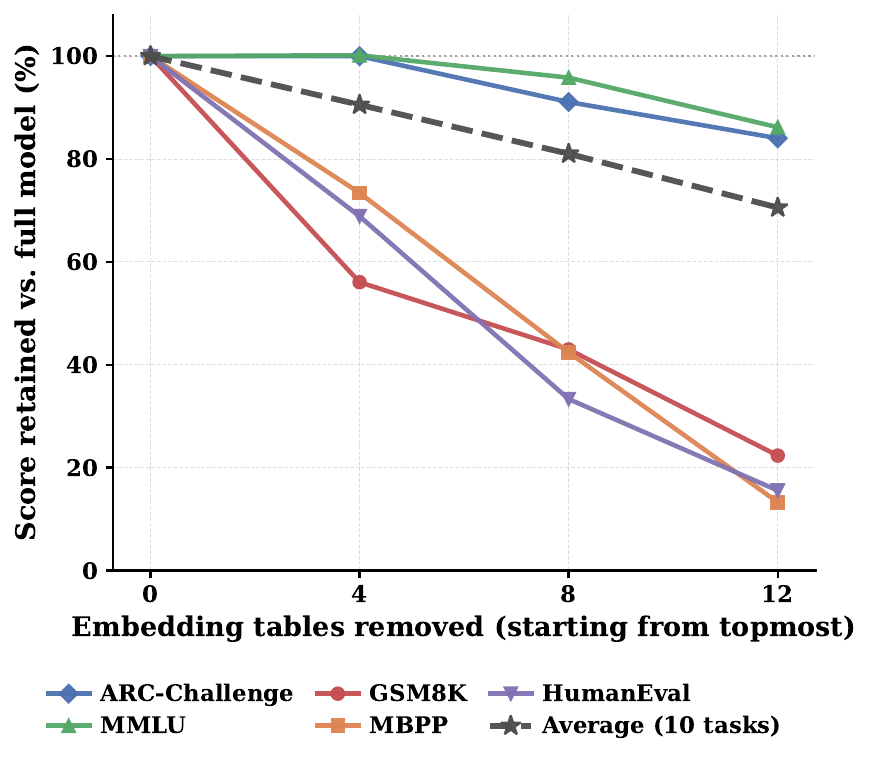}
        \label{fig:layer_ablation}
    }
    \subfloat[Order-statistics intervention]{
        \includegraphics[width=0.32\linewidth]{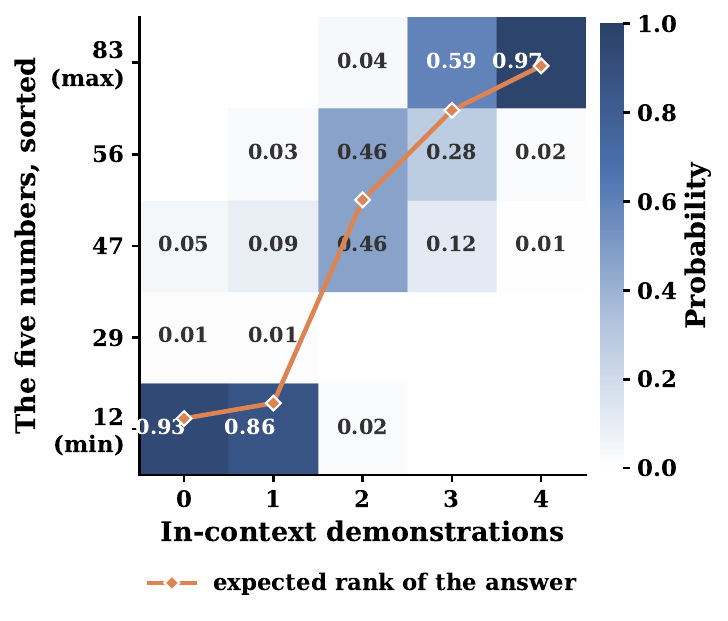}
        \label{fig:maxmin_rank_sweep}
    }
    \subfloat[Code-generation intervention]{
        \includegraphics[width=0.32\linewidth]{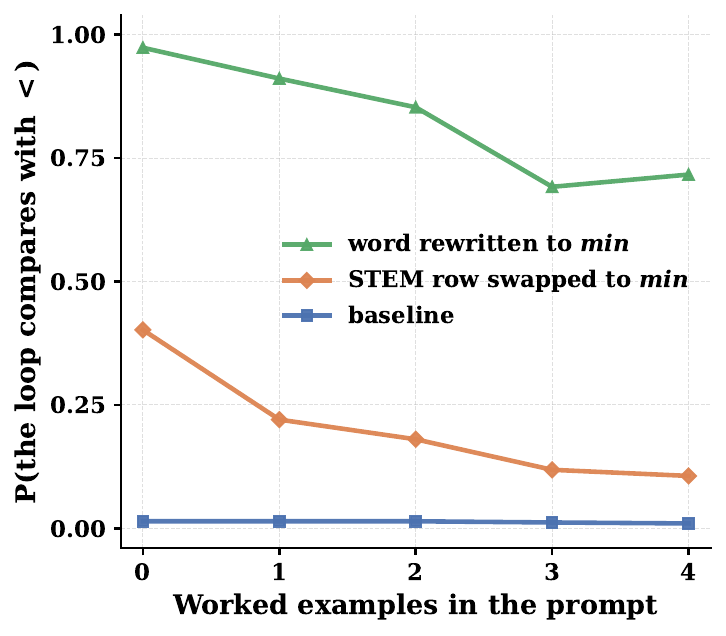}
        \label{fig:code_maxmin_sweep}
    }
        \caption{\textbf{Static memory ablation and controlled embedding interventions.} \textbf{(a)} Benchmark-score retention after removing the last 4, 8, or all 12 embedding tables from trained \staple-12L. \textbf{(b)} Order statistics: with the \stem embedding for `maximum' swapped for that of `minimum', probability over the five sorted candidates (heatmap) and expected answer rank (line) as correct demonstrations are added. \textbf{(c)} Code generation: probability that the generated maximum-finding loop compares with $<$ under the same swap, alongside the unmodified model and the instruction rewritten to `minimum'.}
\end{figure*}

The removal ablations show that \staple relies on its static memory, but not how that memory affects individual predictions. Following \stem~\citep{sadhukhan2026stemscalingtransformersembedding}, we probe this with causal interventions on the embeddings. Because each row of $U_\ell$ is selected only by the input token id, substituting one token's row for another at a selected position alters the memory contribution without changing the input text, so any change in the output is attributable to the embeddings. At the same time, Eq.~(\ref{eq:staple_additive}) scales $U_\ell[t]$ by the context-dependent gate $\mathrm{SiLU}(W^g_\ell x_\ell)$, so context may counteract a conflicting memory signal. We test this on two tasks, finding the maximum of five numbers and writing Python code that finds the maximum of an array. In both, we replace the \stem embedding for `maximum' with that for `minimum' and vary the number of correct demonstrations in the prompt.

\textbf{Order statistics.}
Without few-shot demonstrations, the swap overrides the instruction: the model assigns $0.93$ probability to the minimum (Fig.~\ref{fig:maxmin_rank_sweep}). Each demonstration shifts probability toward larger answers, and with four the model assigns $0.97$ probability to the maximum.

\textbf{Code generation.}
The swap also alters the generated algorithm, since comparing with $<$ rather than $>$ turns the loop into a minimum finder. The probability of $<$ is $0.40$ without demonstrations and falls to $0.11$ with four (Fig.~\ref{fig:code_maxmin_sweep}). For reference, the unmodified model almost never uses $<$, while rewriting the instruction to `minimum' makes $<$ the most likely choice at every demonstration count.

In both tasks, the static embeddings causally change the output, from the selected answer to the program's logic, yet consistent context progressively overrides them. This is consistent with the context-dependent gate controlling how much the token-indexed memory contributes.

We further trace an implicit numerical update through a GSM8K solution in Appendix~\ref{app:gsm8k_memory_intervention} (Fig.~\ref{fig:gsm8k_memory_intervention}). With the question unchanged, a vocabulary-id-based \stem row remapping from \texttt{60} to \texttt{120} shifts the intermediate quantities and final answer to match an explicitly rewritten question, illustrating how a targeted memory intervention can propagate through downstream reasoning.

\section{Related Works}
\paragraph{Static sparse memory.}
Token-indexed memory expands language-model capacity with little additional computation, as explored by Gemma 3n~\citep{gemma3n2024}, \stem~\citep{sadhukhan2026stemscalingtransformersembedding}, and JTok~\citep{yang2026jtoktokenembeddingaxis}. MeKi~\citep{ding2026mekimemorybasedexpertknowledge} adds memory parallel to the FFN, while L$^3$~\citep{tseng2026l3largelookuplayers} contextually aggregates statically selected embeddings. Engram~\citep{cheng2026conditionalmemoryscalablelookup} uses hashed $n$-gram memory with contextual gating; LongCat-Flash-Lite~\citep{liu2026scalingembeddingsoutperformsscaling} also studies $n$-gram embedding scaling, and Qwen3.8-Flash-Next~\citep{qwen2026design,qwen3.8flashnext} demonstrates host-offloaded memory at scale. \staple studies how to make newly added static capacity useful in mature pretrained checkpoints under limited continued training. It retains the pretrained up-projection that \stem replaces and adds zero-initialized token residuals within the FFN. Our architecture and warmup studies examine this adaptation recipe, and our parameter-matched Engram-style baseline compares an alternative memory design under the same upcycling budget.

\paragraph{Memory-based adaptation.}
Learned key--value memory layers provide another form of sparse capacity~\citep{lample2019largememorylayersproduct,berges2024memorylayersscale}. Sparse memory finetuning selectively updates memory slots to acquire knowledge while reducing forgetting~\citep{lin2025continuallearningsparsememory}, and subsequent work retrofits pretrained Transformers with sparse memory and recovery training~\citep{goyal2026improvingsparsememoryfinetuning}. These studies emphasize knowledge acquisition and retention. \staple instead examines how deterministically indexed memory scales broad downstream performance during continued training, combining preservation of pretrained FFN computation with predictable retrieval for training and serving offload.

\paragraph{Sparse upcycling.}
Sparse upcycling reuses pretrained dense weights to initialize MoEs~\citep{komatsuzaki2023sparseupcyclingtrainingmixtureofexperts,he2025upcyclinglargelanguagemodels,vavre2024llama3meetsmoe}. Domain-specialized branches~\citep{sukhbaatar2024branchtrainmixmixingexpertllms}, informed router initialization~\citep{jiang2025improvedsparseupcycling}, and partial expert reinitialization~\citep{nakamura2025dropupcyclingtrainingsparsemixture} address expert specialization. Other studies characterize training-budget tradeoffs~\citep{liew2025scalinglawsupcyclingmoe} and serving overheads~\citep{doubov2024sparseupcyclinginference}. \staple expands token-indexed memory while retaining dense FFN computation, avoiding cloned expert pools and learned routing. Its central challenge is making sparsely and unevenly updated parameters useful within a limited training budget; we study architecture, initialization, and warmup choices that enable this expansion while largely preserving the seed model's throughput.
\section{Conclusion}
We introduced \staple, a simple and sample-efficient method for expanding the parametric capacity of pretrained dense models during mid-training. \staple shows that the system-efficient scaling of static sparsity can translate into practical performance gains even in the upcycling setting. The key idea is to preserve the pretrained model's learned knowledge and to further complement it with scalable memory. Despite introducing a large number of new parameters, \staple remains highly token-efficient and provides consistent gains with increasing memory scale, notably on reasoning-heavy tasks. Our training recipe enables static sparse upcycling of pretrained models, and compared with a serving-efficient MoE upcycling baseline, \staple achieves a higher average score while training faster and decoding 2$\times$ faster with a much smaller GPU memory footprint. An exciting direction for future work is whether jointly scaling static memory with MoE can continue to push the performance--efficiency frontier.

\bibliographystyle{plainnat}
\bibliography{main}
\clearpage
\newpage
\beginappendix
\section{Training hyperparameters}
\label{app:hyperparams}

\begin{table}[h]
  \caption{Training hyperparameters for warmup and mid-training. \emph{STEM/STAPLE tables} are the token-indexed static sparse embeddings; \emph{Engram n-gram tables} are the hashed n-gram embeddings of the Engram-style baseline. During \staple warmup, pretrained up-projections in augmented layers are frozen. The backbone warmup schedule resumes the base model's pretraining cosine schedule at its final checkpoint. MoE expert weights use a separate cosine schedule during warmup and linear decay during mid-training, with a peak learning rate of $2\times10^{-4}$ in both stages.}
  \label{tab:training-hyperparams}
  \centering
  \small
  \setlength{\tabcolsep}{5pt}
  \renewcommand{\arraystretch}{1.15}
  \resizebox{\columnwidth}{!}{%
  \begin{tabular}{llcccccccccc}
    \toprule
    \multicolumn{2}{c}{Parameter group}
    & \multicolumn{4}{c}{Training setup}
    & \multicolumn{6}{c}{Optimization schedule} \\
    \cmidrule(r){1-2}
    \cmidrule(lr){3-6}
    \cmidrule(l){7-12}
    Stage & Parameters
    & Seq.\ Len. & Batch Size & Train Steps & Warmup Steps
    & Resume Step & End Step & LR Scheduler & Peak LR & Min LR Ratio & Weight Decay \\
    \midrule
    \multirow{4}{*}{Warmup}
      & \makecell[l]{STEM/STAPLE\\tables} & 4096 & 512 & 50{,}000 & 500
      & 0 & 50{,}000 & cosine & $4 \times 10^{-4}$ & 0.6 & $1 \times 10^{-4}$ \\
      & \makecell[l]{Engram n-gram\\tables} & 4096 & 512 & 50{,}000 & 500
      & 0 & 50{,}000 & cosine & $4 \times 10^{-4}$ & 0.6 & $1 \times 10^{-4}$ \\
      & Backbone & -- & -- & 50{,}000 & 0
      & 1{,}907{,}359 & 2{,}384{,}186 & cosine & $4 \times 10^{-4}$ & 0.1 & $1 \times 10^{-1}$ \\
      & MoE expert weights & -- & -- & 50{,}000 & 500
      & 0 & 50{,}000 & cosine & $2 \times 10^{-4}$ & 0.91 & $1 \times 10^{-1}$ \\
    \midrule
    \multirow{4}{*}{Mid-training}
      & \makecell[l]{STEM/STAPLE\\tables} & 4096 & 512 & 50{,}000 & 0
      & 0 & 50{,}000 & linear & $3 \times 10^{-4}$ & 0 & $1 \times 10^{-4}$ \\
      & \makecell[l]{Engram n-gram\\tables} & 4096 & 512 & 50{,}000 & 0
      & 0 & 50{,}000 & linear & $3 \times 10^{-4}$ & 0 & $1 \times 10^{-4}$ \\
      & Backbone & -- & -- & 50{,}000 & 0
      & 0 & 50{,}000 & linear & $7.44 \times 10^{-5}$ & 0 & $1 \times 10^{-1}$ \\
      & MoE expert weights & -- & -- & 50{,}000 & 0
      & 0 & 50{,}000 & linear & $2 \times 10^{-4}$ & 0 & $1 \times 10^{-1}$ \\
    \bottomrule
  \end{tabular}%
  }
\end{table}

\section{Serving latency and throughput}
\label{app:serving}

Table~\ref{tab: inference table} reports the full serving measurements summarized in Fig.~\ref{fig:decode_throughput}, including batch-size-1 decode and prefill latency and throughput.

\begin{table*}[h]
\centering
\caption{\textbf{Serving efficiency on one NVIDIA B200.} All measurements use context length 256; latency and throughput are reported in ms and tok/s. HBM memory includes resident weights and GPU caches. \staple-12L offloads 12 embedding tables (19.73\,GB in BF16 precision) to CPU DRAM with a 1\,GB GPU cache. MoE denotes the upcycled MoE baseline of Sec.~\ref{sec:setup} (eight interleaved MoE layers, each with eight full-width experts and top-1 routing), which keeps all experts in HBM and incurs routing and token-reordering overhead.}
\label{tab: inference table}
\small
\setlength{\tabcolsep}{3.8pt}
\renewcommand{\arraystretch}{1.12}
\resizebox{\textwidth}{!}{%
\begin{tabular}{@{}lcccccc@{}}
\toprule
\textbf{Architecture}
& \makecell{\textbf{HBM model}\\\textbf{memory}}
& \makecell{\textbf{Linear FLOPs}\\\textbf{/ token}}
& \makecell{\textbf{Decode B1}\\\textbf{ms / tok/s}}
& \makecell{\textbf{Decode B64}\\\textbf{ms / tok/s}}
& \makecell{\textbf{Prefill B1}\\\textbf{ms / tok/s}}
& \makecell{\textbf{Prefill B64}\\\textbf{ms / tok/s}} \\
\midrule
Dense
& 2.97 GB GPU weights
& $\sim$2.56 G
& 1.224 / 817
& 1.817 / 35.2K
& 1.419 / 180.5K
& 34.184 / 479.3K \\
\staple-12L
& 2.97 GB GPU weights + 1 GB cache
& $\sim$2.56 G
& 1.237 / 809
& 1.845 / 34.7K
& 1.524 / 168.0K
& 34.678 / 472.5K \\
\stem-12L
& 2.57 GB GPU weights + 1 GB cache
& $\sim$2.16 G
& 1.151 / 869
& 1.671 / 38.3K
& 1.413 / 181.1K
& 27.580 / 594.1K \\
MoE (8 experts, top-1)
& 8.61 GB GPU weights, all 8 experts
& $\sim$2.56 G + routing
& 1.951 / 513
& 3.761 / 17.0K
& 2.975 / 86.1K
& 36.179 / 452.9K \\
\bottomrule
\end{tabular}%
}
\end{table*}

\paragraph{Energy measurements and host-cost estimates.}
\label{app:gpu_energy}
We use an NVIDIA DGX B200 system (board \texttt{DGXB200}, version A.5) with dual Intel Xeon Platinum 8570 processors and PCIe Gen5 $\times$16 CPU--GPU links. In a separate energy-profiling run on one otherwise idle B200 GPU, we measure decode at batch size 64 and context length 256 using Zeus~\citep{zeus-nsdi23}, with three repeats per model agreeing within 1.5\%. GPU energy per generated token is 18.69\,mJ for dense, 18.92\,mJ for \staple-12L, and 45.37\,mJ for the MoE baseline: \staple adds approximately 1.2\% over dense and uses 58.3\% less than MoE. Activity-based estimates complement these GPU measurements: multiplying measured host-memory and PCIe traffic by conservative per-bit energy coefficients places \staple's additional host-memory and transfer energy below 0.2\% of its GPU energy. The CPU launch-thread energy is estimated at approximately 1\% of GPU energy for all three models, leaving the comparison essentially unchanged.

\section{Implementation Details}
\label{sec: implementation}

We keep the large \stem tables in CPU DRAM and move only the rows required by the current workload to the GPU. Inference uses a small GPU-resident LFU cache, whereas training materializes a compact buffer over the distinct token set described in Sec.~\ref{sec: offload}.

\subsection{Inference-time embedding offload}

Prefill and decode require different host-to-device embedding retrieval paths because of the difference in their prefetching workloads. Table~\ref{tab:inference_impl_summary} summarizes the implementation; in both cases, retrieval is overlapped with backbone execution. Our serving stack additionally uses CUDA graphs~\citep{gray2019cudagraphs}, a fused in-place KV-cache append kernel, and FlashInfer~\citep{ye2025flashinfer} for paged decode attention.

\begin{table}[h]
\centering
\footnotesize
\caption{Inference-time retrieval paths for host-offloaded \stem tables. Prefill favors bulk deduplicated transfers, whereas decode uses GPU-driven fine-grained lookup to minimize exposed latency.}
\label{tab:inference_impl_summary}
\setlength{\tabcolsep}{4pt}
\renewcommand{\arraystretch}{1.10}
\begin{tabular}{@{}p{0.10\linewidth}p{0.40\linewidth}p{0.42\linewidth}@{}}
\toprule
\textbf{Phase} & \textbf{Retrieval path} & \textbf{Overlap / execution} \\
\midrule
\textbf{Prefill} &
\begin{itemize}[leftmargin=*,nosep,topsep=0pt,partopsep=0pt]
    \item Deduplicate prompt tokens once.
    \item Probe the GPU LFU cache and gather only cache-miss rows from CPU.
    \item Pack misses into pinned host buffers and issue bulk asynchronous H2D copies.
\end{itemize}
&
\begin{itemize}[leftmargin=*,nosep,topsep=0pt,partopsep=0pt]
    \item Reuse the same inverse token map across all \staple layers.
    \item Prefetch later-layer rows while earlier transformer layers execute.
    \item Pipeline successive requests/chunks and fuse compact-row reconstruction with the FFN using \texttt{torch.compile}.
\end{itemize}
\\
\addlinespace[4pt]
\textbf{Decode} &
\begin{itemize}[leftmargin=*,nosep,topsep=0pt,partopsep=0pt]
    \item Store every four adjacent \staple tables in a token-major CPU layout.
    \item Launch a GPU lookup kernel that probes the LFU cache.
    \item On a miss, read the row directly from CUDA-registered host memory; no CPU gather or explicit H2D copy is required.
\end{itemize}
&
\begin{itemize}[leftmargin=*,nosep,topsep=0pt,partopsep=0pt]
    \item Capture grouped lookups and model execution in a multi-stream CUDA graph.
    \item Launch later lookup groups on a side stream while earlier layers compute.
    \item Synchronize only when a group is first consumed (layers 1, 5, and 9 for \staple-12L).
\end{itemize}
\\
\bottomrule
\end{tabular}
\end{table}

For very short prefills, where CPU-gather setup dominates, we use the same GPU-driven mapped-host path as decode. This avoids CPU gathers and explicit H2D copies on the latency-sensitive path.

\subsection{Training-time embedding offload}
\label{sec:training_offload_impl}

Algorithm~\ref{alg:training_offload} shows the training pipeline. The token union and inverse map are computed once per step and reused by every \stem layer; full tables and fp32 AdamW moments remain in host memory.

\begin{algorithm}[!htbp]
\footnotesize
\DontPrintSemicolon
\caption{Training step with host-offloaded \stem tables.}
\label{alg:training_offload}
\KwIn{batch token ids $\mathcal{B}$; host tables/states $\{E_\ell,m_\ell,v_\ell\}$}
$(\mathcal{U},\pi) \leftarrow \mathrm{unique}(\mathcal{B})$\tcp*[r]{once per step}

\For{$\ell \in \mathcal{L}_{\mathrm{STEM}}$ in forward order}{
    enqueue host gather $R_\ell \leftarrow E_\ell[\mathcal{U}]$ into pinned memory\;
    enqueue async H2D copy $R_\ell \rightarrow \widehat{R}_\ell$ and record ready event\;
}

run forward; at each \stem layer, wait for its ready event and use $\widehat{R}_\ell[\pi]$\;

run backward\;
\ForEach{\stem layer $\ell$ when compact gradient $G_\ell$ becomes ready}{
    enqueue async D2H copy of $G_\ell$ into the alternate staging buffer\;
    update $(E_\ell,m_\ell,v_\ell)$ with row-restricted AdamW on rows $\mathcal{U}$\;
}
\end{algorithm}

The compact $U\times d_{\mathrm{ff}}$ row buffer is the autograd leaf, so backward never allocates a dense $V\times d_{\mathrm{ff}}$ gradient. Gradient staging is double buffered, while H2D and D2H transfers use separate streams. We issue next-step gathers in forward consumption order rather than inheriting backward order, which bounds the exposed pipeline-fill latency for the first augmented layer.

Finally, host-memory placement is important for sustaining bandwidth: table allocations are interleaved across NUMA memory controllers and optionally backed by huge pages, while the OpenMP worker pool uses a passive idle policy to avoid competing with the GPU-launch thread. Latency-critical CUDA-event waits are instead polled by the dedicated host worker.

\section{Tracing an implicit numerical update through downstream reasoning}
\label{app:gsm8k_memory_intervention}

The ablations establish reliance on static memory, and Section~\ref{subsec:interpretability} shows that individual embeddings can steer predictions. Here we trace how an intervention propagates through a multi-step GSM8K solution.

\paragraph{Setup.}
We use four-shot chain-of-thought context on items the unmodified model solves correctly, separately from the five-shot benchmark evaluation in Section~\ref{sec:setup}. Unlike the position-specific probes, this intervention is keyed to a vocabulary id: the \stem row for \texttt{60} is remapped to the row for \texttt{120} in the intervened tables, leaving the input text and dense weights unchanged. The remapping applies wherever the source id occurs, including in the model's own continuation.

\begin{figure}[H]
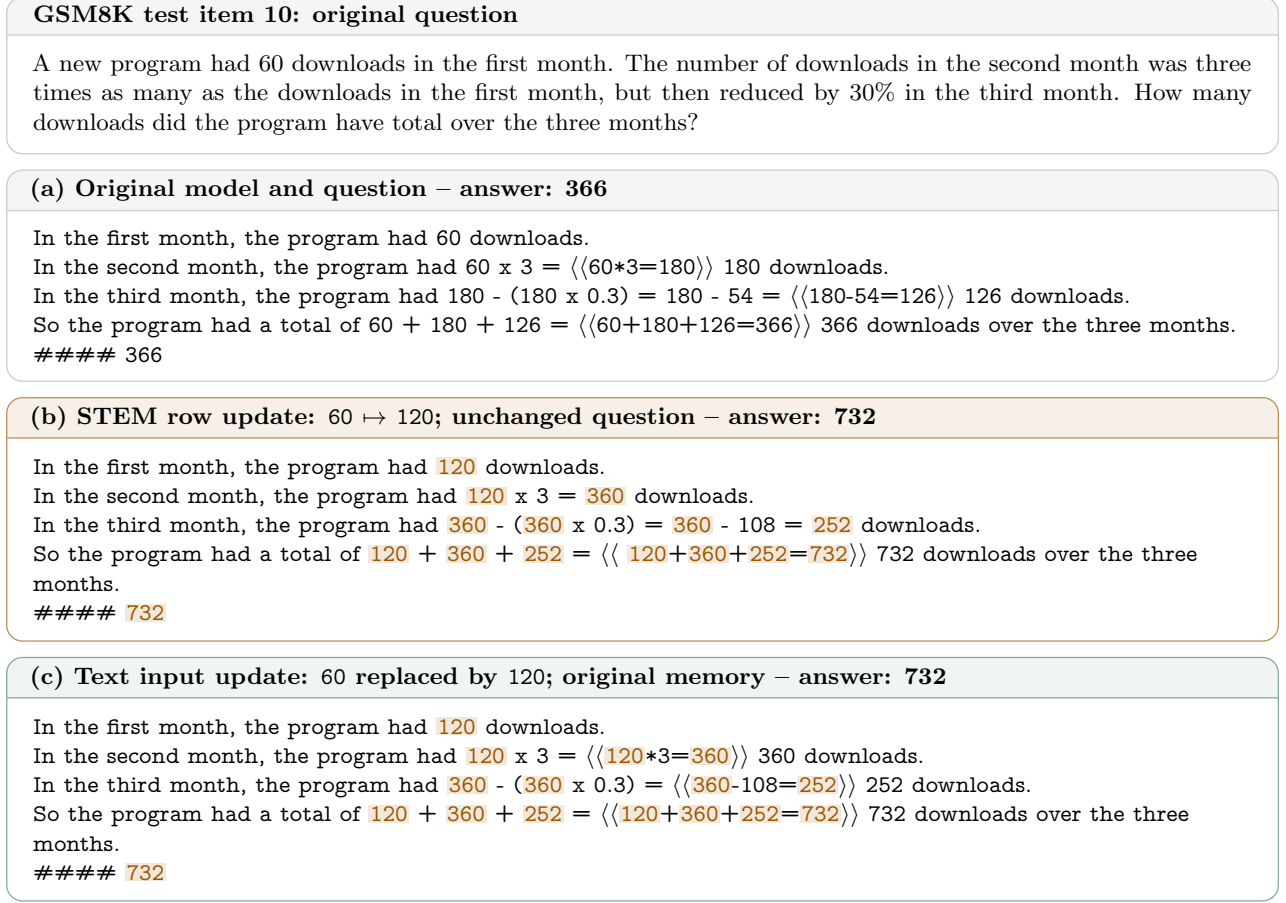

\centering
\begingroup
\definecolor{gsmaccent}{HTML}{9A5700}
\definecolor{gsmcontrol}{HTML}{28645B}
\newcommand{\gsmchanged}[1]{{\setlength{\fboxsep}{0.6pt}\colorbox{gsmaccent!14}{\textcolor{gsmaccent}{\ttfamily\bfseries #1}}}}
\tcbset{gsmcard/.style={enhanced,boxrule=0.5pt,arc=2mm,
  left=2.4mm,right=2.4mm,top=1.5mm,bottom=1.5mm,
  fonttitle=\small\bfseries,coltitle=black,
  colback=white,colframe=black!20,colbacktitle=black!4,
  before skip=0pt,after skip=0pt}}
\begin{tcolorbox}[gsmcard,title={GSM8K test item 10: original question}]
\small
A new program had 60 downloads in the first month. The number of downloads in the second month was three times as many as the downloads in the first month, but then reduced by 30\% in the third month. How many downloads did the program have total over the three months?
\end{tcolorbox}
\vspace{2mm}

\begin{tcolorbox}[gsmcard,title={(a) Original model and question -- answer: 366}]
\small\ttfamily\raggedright
In the first month, the program had 60 downloads.\\
In the second month, the program had 60 x 3 = $\langle\langle$60*3=180$\rangle\rangle$ 180 downloads.\\
In the third month, the program had 180 - (180 x 0.3) = 180 - 54 = $\langle\langle$180-54=126$\rangle\rangle$ 126 downloads.\\
So the program had a total of 60 + 180 + 126 = $\langle\langle$60+180+126=366$\rangle\rangle$ 366 downloads over the three months.\\
\#\#\#\# 366
\end{tcolorbox}
\vspace{2mm}

\begin{tcolorbox}[gsmcard,
  title={(b) STEM row update: \texttt{60} $\mapsto$ \texttt{120}; unchanged question -- answer: 732},
  colframe=gsmaccent!65,colbacktitle=gsmaccent!9]
\small\ttfamily\raggedright
In the first month, the program had \gsmchanged{120} downloads.\\
In the second month, the program had \gsmchanged{120} x 3 = \gsmchanged{360} downloads.\\
In the third month, the program had \gsmchanged{360} - (\gsmchanged{360} x 0.3) = \gsmchanged{360} - 108 = \gsmchanged{252} downloads.\\
So the program had a total of \gsmchanged{120} + \gsmchanged{360} + \gsmchanged{252} = $\langle\langle$ \gsmchanged{120}+\gsmchanged{360}+\gsmchanged{252}=\gsmchanged{732}$\rangle\rangle$ 732 downloads over the three months.\\
\#\#\#\# \gsmchanged{732}
\end{tcolorbox}
\vspace{2mm}

\begin{tcolorbox}[gsmcard,
  title={(c) Text input update: \texttt{60} replaced by \texttt{120}; original memory -- answer: 732},
  colframe=gsmcontrol!55,colbacktitle=gsmcontrol!7]
\small\ttfamily\raggedright
In the first month, the program had \gsmchanged{120} downloads.\\
In the second month, the program had \gsmchanged{120} x 3 = $\langle\langle$\gsmchanged{120}*3=\gsmchanged{360}$\rangle\rangle$ 360 downloads.\\
In the third month, the program had \gsmchanged{360} - (\gsmchanged{360} x 0.3) = $\langle\langle$\gsmchanged{360}-108=\gsmchanged{252}$\rangle\rangle$ 252 downloads.\\
So the program had a total of \gsmchanged{120} + \gsmchanged{360} + \gsmchanged{252} = $\langle\langle$\gsmchanged{120}+\gsmchanged{360}+\gsmchanged{252}=\gsmchanged{732}$\rangle\rangle$ 732 downloads over the three months.\\
\#\#\#\# \gsmchanged{732}
\end{tcolorbox}
\endgroup
\vspace{2mm}
\caption{\textbf{A token-indexed memory intervention propagates through a GSM8K solution.}
Full output text for the original model, the \stem row update, and the rewritten-question control, with line wrapping adjusted for readability. Highlights mark the critical numerical changes. With the question unchanged, the row update produces the same intermediate quantities and final answer as explicitly replacing 60 with 120 in the question.}
\label{fig:gsm8k_memory_intervention}
\end{figure}

\paragraph{Observation and interpretation.}
The row update changes the monthly quantities from 60, 180, and 126 to 120, 360, and 252, and the total from 366 to 732. Agreement with the rewritten-question control supports interpreting this as a change in the effective numerical premise used by subsequent computation. The example links a token-addressable memory intervention to coherent downstream reasoning, but does not establish general edit reliability or an isolated fact representation. Since remapping applies to all source-id occurrences, it does not separate prompt-processing effects from generated-token feedback.

\end{document}